\documentclass[ijds,nonblindrev]{informs-ijds}

\OneAndAHalfSpacedXI

\usepackage{natbib}
 \bibpunct[, ]{(}{)}{,}{a}{}{,}%
 \def\bibfont{\small}%
\TheoremsNumberedThrough     
\ECRepeatTheorems

\EquationsNumberedThrough    

\MANUSCRIPTNO{IJDS-xxxx-xxxx.xx}

\usepackage{tikz, pgfplots}
\usepackage[font=small]{subcaption}
\pgfplotsset{compat=1.17}

\begin{document}



\RUNTITLE{Fern\'{a}ndez-Lor\'{i}a and Provost}

\TITLE{Causal Decision Making and Causal Effect Estimation Are Not the Same... and Why It Matters}

\ARTICLEAUTHORS{%
\AUTHOR{\textit{To appear in the inaugural issue of the INFORMS Journal of Data Science.}}
\AUTHOR{Carlos Fern\'{a}ndez-Lor\'{i}a}
\AFF{HKUST Business School, \EMAIL{imcarlos@ust.hk}} 
\AUTHOR{Foster Provost}
\AFF{NYU Stern School of Business and Compass Inc. \EMAIL{fprovost@stern.nyu.edu}}

} 

\ABSTRACT{%
Causal decision making (CDM) at scale has become a routine part of business, and increasingly CDM is based on statistical models and machine learning algorithms. Businesses algorithmically target offers, incentives, and recommendations to affect consumer behavior. Recently, we have seen an acceleration of research related to CDM and causal effect estimation (CEE) using machine-learned models. This article highlights an important perspective: CDM is not the same as CEE, and counterintuitively, accurate CEE is not necessary for accurate CDM. Our experience is that this is not well understood by practitioners or most researchers. Technically, the estimand of interest is different, and this has important implications both for modeling and for the use of statistical models for CDM. We draw on recent research to highlight three implications. (1) We should consider carefully the objective function of the causal machine learning, and if possible, we should optimize for accurate ``treatment assignment" rather than for accurate effect-size estimation. (2) Confounding does not have the same effect on CDM as it does on CEE. The upshot here is that for supporting CDM it may be just as good or even better to learn with confounded data as with unconfounded data. Finally, (3) causal statistical modeling may not be necessary at all to support CDM because a proxy target for statistical modeling might do as well or better. This third observation helps to explain at least one broad common CDM practice that seems ``wrong" at first blush---the widespread use of non-causal models for targeting interventions. The last two implications are particularly important in practice, as acquiring (unconfounded) data on both ``sides" of the counterfactual for modeling can be quite costly and often impracticable. These observations open substantial research ground. We hope to facilitate research in this area by pointing to related articles from multiple contributing fields, including two dozen articles published the last three to four years.

}%



\maketitle

%


\section{Introduction}

Causal decision making at scale has become a routine part of business.  Firms build statistical models using machine learning algorithms to support decisions about targeting all manner of interventions.  The most common are offers, advertisements, retention incentives, and recommendations made to customers and prospective customers. The goal of these interventions is to affect consumer behavior, for example to increase purchasing or to keep a customer from churning.  
When we talk about causal decision making in this article, we specifically mean deciding whether or not to ``treat" a particular individual with a particular intervention.
Making explicit the causal nature of such decision making has lagged its implementation in practice. Today, data-savvy firms routinely conduct causal \textit{evaluations} of their targeting decisions, for example, conducting A/B tests when making offers or targeting advertisements.  However, very few firms are actually building causal models to support their causal decision making.  

The academic community has, of course, taken on (conditional) causal estimation, and over the past 3--4 years, we have seen an acceleration of research on the topic.\footnote{We hope that one contribution of this article is the collection of recent articles on the topic we point to.}  Much of this work has focused on machine learning methods for individual causal effect estimation~\citep{dorie2019automated}.  The main motivation behind several of these methods is causal decision making~\citep{athey2017state,athey2019machine,mcfowland2021prescriptive}.

This article highlights an important perspective: \textbf{Causal decision making (CDM)} is not the same as \textbf{causal effect estimation (CEE)}.  Counterintuitively, accurate CEE is not necessary for accurately assigning treatments to maximize causal effect.  Our experience is that this is not well understood by practitioners nor by most data science researchers---including some who routinely build models for estimating causal effects.  Very recent research has started to examine this question, but much fertile ground for new research remains due to the limited attention this perspective has received.  We will present three main implications for research and practice to highlight how this perspective can be enlightening and impactful, but many other---possibly even more important---implications might be revealed by future research.

The fundamental problem in many predictive tasks is often assessing whether an intervention will have an effect.  For example, when attempting to stop customers from leaving, the goal is not to target the customers most likely to leave, but instead to target the customers whom the incentive will cause to stay. For many advertising settings, the advertiser's goal is not simply to target ads to people who will purchase after seeing the ad, but to target people the ad will cause to purchase. Generally, instead of simply predicting the likelihood of a positive outcome, for tasks such as these we ideally would like to predict whether a particular individual's outcome can be improved by means of an intervention or treatment (i.e., the ad or the retention incentive in the examples). 

This type of causal decision making is an instance of the problem of treatment assignment~\citep{manski2004statistical}. Each possible course of action corresponds to a different ``treatment'' (such as ``target'' vs. ``not target''), and ideally, each individual is assigned to the treatment associated with the most beneficial outcome (e.g., the one with the highest profit). Treatment assignment policies may be estimated from data using statistical modeling, allowing decision makers to map individuals to the best treatment according to their characteristics (e.g., preferences, behaviors, history with the firm). There are many different ways in which one could proceed, and various methods for the estimation of treatment assignment policies from data have been proposed across different fields, including econometrics~\citep{bhattacharya2012inferring}, data mining~\citep{radcliffe2011real}, and machine learning in the contextual bandit setting~\citep{li2010contextual}. 

One approach for causal decision making that is becoming popular in the research literature is to use machine-learned models to estimate causal (treatment) effects at the individual level, and then use the models to make intervention decisions automatically~\citep{olaya2020survey}. For instance, in cases where there is a limited advertising budget, causal effect models may be used to predict the expected effect of a promotion on each potential customer and subsequently target those individuals with the largest predicted effect. 

However, causal decision making (treatment assignment) and causal effect estimation are not the same. Going back to the advertising example, suppose a firm is willing to send an offer to customers for whom it increases the probability of purchasing by at least 1\%. In this case, precisely estimating individual causal effects is desirable but not necessary; the only concern is identifying those individuals for whom the effect is greater than 1\%. Importantly, overestimating (underestimating) the effect has no bearing on decision making when the focal individuals have an effect greater (smaller) than 1\%.  This is very similar to the distinction in non-causal predictive modeling between regression or probability estimation and classification, and we will take advantage of that analogy as we proceed.

The main distinction between causal decision making and causal effect estimation is their estimand of interest. The main object of study in (heterogeneous) causal effect estimation is understanding treatment effects in heterogeneous subpopulations.  On the other hand, causal effect estimates are important for causal decision making only to the extent that they help to discriminate individuals according to preferred treatments.\footnote{Also compare this with distinctions between explanatory modeling and predictive modeling~\citep{shmueli2010explain}.}

This paper draws on recent research to highlight three implications of this distinction.  

\begin{enumerate}
    \item We should consider carefully the objective function of the causal machine learning algorithm, and if possible, it should optimize for accurate treatment assignment rather than for accurate effect estimation. A CDM practitioner reaching for a ``causal" machine learning algorithm geared toward accurately estimating conditional causal effects (as most are) may be making a mistake.  
    
    \item Confounding does not have the same effect on CDM as it does on CEE.  When supporting CDM, it may be as good or even better to learn with confounded data---such as data suffering from uncorrectable selection bias. Acquiring (unconfounded) data on both ``sides" of the counterfactual for modeling can be quite costly and often impracticable.  Furthermore, even when we have invested in unconfounded data---for example, via a controlled experiment---often we have much more confounded data, so there is a causal bias/variance tradeoff to be considered.
    
    \item Causal statistical modeling may not be necessary at all to support CDM, because there may be (and perhaps often is) a proxy target for statistical modeling that can do as well or better.  This third implication helps to explain at least one broad common CDM practice that seems ``wrong" at first blush---the widespread use of non-causal models for targeting interventions.  

\end{enumerate}

Overall, this paper develops the perspective that what might traditionally be considered ``good'' estimates of causal effects are not necessary to make good causal decisions. The three implications above are quite important in practice, because acquiring data to estimate causal effects accurately is often complicated and expensive. Empirically, we see that results can be considerably better when modeling intervention decisions rather than causal effects.  On the research side, as mentioned above, taking this perspective reveals substantial room for novel research.

\section{Distinguishing Between Effect Estimation and Decision Making}

We use the potential outcomes framework~\citep{rubin1974estimating} to define causal effects. The framework defines interventions in terms of treatment alternatives (treatment ``levels"), which may range from two (e.g., to represent absence or presence of a treatment) to many more (e.g., to represent treatment doses). The framework also assumes the existence of one \textbf{potential outcome} associated with each treatment alternative for each individual, which represents the value that the outcome of interest would take if the individual were to be exposed to the associated treatment. \textbf{Causal effects} are defined in terms of the difference between two potential outcomes. Our subsequent discussions and examples focus on cases with only two intervention alternatives, but most of the ideas and insights also apply to cases with more. 

The main difference between causal effect estimation and estimating the best treatment assignment is the estimand of interest. Formally, let $Y(i)\in \mathbb{R}$ be the potential outcome associated with treatment level $i$. The estimand in (heterogeneous) \textbf{causal effect estimation (CEE)} is:
\begin{equation}\label{eq:effect_estimand}
    \tau(x)=E[Y(1)-Y(0)|X=x],
\end{equation}
which corresponds to the \textbf{conditional average treatment effect (CATE)} given the feature vector $X$ used to characterize individuals. Several studies have proposed the use of machine learning methods for CATE estimation \citep[see][for a survey]{dorie2019automated}. Popular methods include Bayesian additive regression trees~\citep{hill2011bayesian}, causal random forests~\citep{wager2018estimation}, and regularized causal support vector machines~\citep{imai2013estimating}. 

The counterpart in \textbf{causal decision making (CDM)}, or treatment assignment, is:
\begin{equation}\label{eq:assignment_estimand}
    a^*(x)= \textbf{1}(\tau(x)>0), 
\end{equation}
which corresponds to the treatment level (or action) that maximizes the expected value of the outcome of interest given the feature vector $X$.\footnote{The threshold for the decision could be a value other than zero, for example when treatment costs and benefits are taken into account; see the discussion below.} Why are causal effect models so often considered the natural choice for causal decision making?  Because Equation~\ref{eq:assignment_estimand} is defined in terms of Equation~\ref{eq:effect_estimand}: individuals should be assigned to treatment level $a^*=1$ when their CATE is positive, and to treatment level $a^*=0$ otherwise. 

However, models provide only \emph{estimates} of causal effects, and estimates incorporate errors.  It turns out that the sorts of errors incorporated by the estimation procedures are critical.
Moreover, as we discuss in detail, accurate causal effect estimation is not necessary to make good intervention decisions, and more accurate estimates of causal effects may not imply better decisions.

A second important distinction between causal effect estimation and causal decision making is that the outcome of interest may differ between the two. For example, in advertising settings, machine learning could be used to predict treatment effects on purchases with the intent of making treatment decisions that maximize profits. However, the individuals for whom treatments are most effective---in terms of increasing the likelihood of purchase---are not necessarily the ones for whom the promotions are most profitable~\citep{miller2020personalized,lemmens2020managing}, so making decisions based on CATEs may lead to sub-optimal treatment assignments in these cases. On the other hand, when the outcomes of interest in the two tasks match (e.g., when CATEs are defined in terms of profits rather than purchases), making decisions based on the true CATEs as defined in Equation~\ref{eq:assignment_estimand} once again leads to optimal treatment assignments. The problem with estimation errors remains. 

A third distinction is that CDM may involve additional constraints that are critical for the optimization problem at hand. For example, targeting the individuals with the largest CATEs may not be optimal when there are budget constraints and costs vary across individuals. One way to proceed in such cases is to use machine learning to predict CATEs and then use those predictions to solve a downstream optimization problem that incorporates the business constraints~\citep{mcfowland2021prescriptive}. However, recent research suggests that decisions can significantly improve when both the business objective and the business constraints are incorporated as part of the machine learning~\citep{elmachtoub2021smart,elmachtoub2020decision}.

More generally, this work contributes to the burgeoning stream of research that recognizes that model learning is often simply a means to a higher-level goal. In our case, learning a CEE model is often only a step towards CDM, and as we demonstrate next, sometimes circumventing this step and directly optimizing for decision making can lead to better results. Similarly, another common goal when learning CEE models is to characterize and identify subpopulations affected by the treatment~\citep{mcfowland2018efficient}. Once again, the end goal for this application is subtly different from CEE or CDM because the estimand of interest corresponds to the identification of the feature values of subpopulations affected by the treatment, and results can be substantially better when algorithms are designed with that specific purpose in mind~\citep{mcfowland2018efficient}. In the context of marketing~\citep{lemmens2020managing} and operations~\citep{elmachtoub2021smart}, researchers have also reported substantially better results when models are directly optimized to minimize decision-making errors instead of prediction errors. Thus, we hope this perspectives paper will help emphasize the importance of accounting for how model predictions will be used when building data-driven solutions.

Another task closely related to CDM as defined in this paper is the contextual multi-armed bandit (MAB) problem~\citep{slivkins2019introduction}, which consists of estimating the arm (treatment level) with the largest reward (potential outcome) given some context (feature vector). However, an important distinction with respect to our setting is that the goal in bandit problems is to learn a CDM model while actively making treatment assignment decisions for incoming subjects. Therefore, an exploration/exploitation dilemma plays an important role in the decision making procedure, whereas in our case, the decision-maker cannot re-estimate the treatment assignment policy after making each decision.\footnote{Few firms have the ability to deploy full-blown online machine learning systems that can manage the exploration/exploitation trade-off dynamically.  It is much more common to deploy the learned models/prediction systems than the machine learning systems themselves.} Thus, our setting is commonly referred to as ``offline learning'' by the MAB community~\citep{beygelzimer2009offset}. 
Nonetheless, we hope several of the ideas discussed here will also be revealing to researchers in this community.  Reward regression is a common solution for deciding which arm to choose. However, the perspective of this paper is that reward regression and estimating the arm with the largest reward are not the same task. So, the implications discussed here could also help inform future methodological developments in the MAB community.

\section{Why It Matters}

Let's consider three important practical implications that result from causal decision making and causal effect estimation having different estimands of interest. 

\subsection{Choosing the Objective Function for Learning}

The first implication is that a machine learning model optimized for treatment assignment can lead to better decisions than a machine learning model optimized for causal effect estimation~\citep{fernandez2020comparison}. Machine learning procedures that estimate causal effects are designed to segment individuals according to how their effects vary, but doing so is a more general problem than segmenting individuals according to how they should be treated. By focusing on a specific task (treatment assignment) rather than on a more general task (estimating causal effects), machine learning procedures can better exploit the statistical power available.

Formally, the (sometimes unstated) goal of machine learning methods that estimate causal effects is to optimize the (possibly regularized) \textbf{mean squared error (MSE)} for treatment effects:
\begin{equation}\label{eq:mse}
\text{MSE}(\hat{\tau}) = E[((Y(1)-Y(0))-\hat{\tau}(X))^2],
\end{equation}
which is minimized when $\hat{\tau}=\tau$ (see Equation~\ref{eq:effect_estimand}). The main challenge in causal inference is that only one potential outcome is observed for any given individual, so Equation~\ref{eq:mse} cannot be calculated directly because $Y(1)-Y(0)$ is not observable. Fortunately, alternative formulations may be used to estimate the treatment effect MSE from data~\citep{schuler2018comparison}, allowing one to compare (and optimize) models based on how good they are at predicting causal effects. 

\begin{figure*}
    \centering
     \begin{tikzpicture}
        \linespread{0.5}
        \begin{axis}[
            ylabel = {Causal effect},
            ymax=4,
            ymin=-1,
            xmax= 2.5,
            xmin = 0.5,
            xticklabels={Model 1, Model 2}, xtick={1,2},
            axis y line*=left,
            axis x line*=bottom]
        \addplot[only marks, mark=*,red] plot coordinates {
            (1, 3.5)
            (2, -0.5)};
        \addlegendentry{~Predictions ($\hat{\tau}$)}
        \addplot[only marks, mark=diamond*,blue, mark size=3pt] plot coordinates {
            (1, 0.5)
            (2, 0.5)};
        \addlegendentry{~True effect ($\tau$)}
        \addplot[no marks,dashed] plot coordinates {
            (1, 3.5)
            (1, 0.5)};
        \addlegendentry{~Prediction error}
        \addplot[style={thick}] coordinates {(0,0) (2.5,0)};
        \addlegendentry{~Decision boundary}
        \addplot[no marks,dashed] plot coordinates {
            (2, 0.5)
            (2, -0.5)};
    \end{axis}
    \end{tikzpicture}
    \caption{Comparison of the causal effect predictions made by two models. Model 1 has a larger prediction error than Model 2, but a smaller decision-making error because it predicts the sign of the causal effect correctly.}
    \label{fig:effect}
\end{figure*}
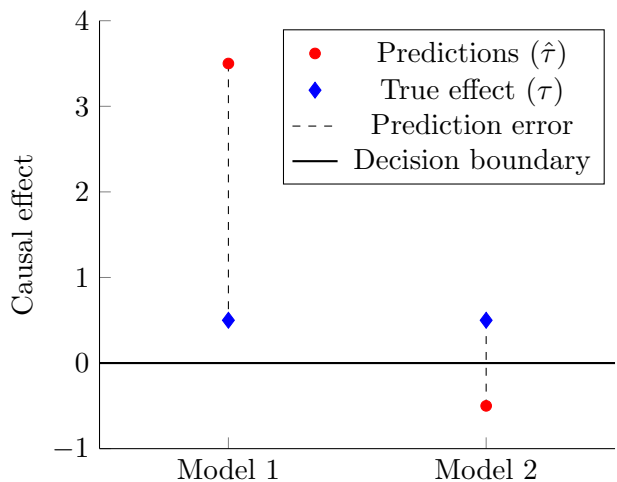

However, optimizing causal effect predictions (by minimizing Equation~\ref{eq:mse}) is not the same as optimizing treatment assignments. We illustrate this in Figure~\ref{fig:effect}, which compares the causal effect predictions made by two different models for a single individual, one with a high effect-prediction error (Model 1) and another with a low effect-prediction error (Model 2). The blue (dark) diamonds correspond to the ``true" CATE for the individual (they are the same for both models), whereas the red dots correspond to the predictions made by the models. A larger distance between the blue diamonds and the red dots (represented by dashed lines) implies that the model has a larger MSE and therefore makes worse causal effect predictions.

In this example, the true CATE is greater than 0 (and hence above the decision boundary), which implies that this individual should be treated. Therefore, if $\hat{\tau}(x) > 0$, $\hat{\tau}$ leads to an optimal treatment assignment.  Consider the zero line to be the causal effect of the ``don't treat" decision.\footnote{This allows what follows to apply when more than only one possible treatment exists.}  Figure~\ref{fig:effect} shows that the model with the larger prediction error makes the better treatment assignment because the rank ordering of the treatment alternatives (``treat'' and ``not treat'') when using the predicted effect is the same as the rank ordering when using the CATE. The second model makes a worse treatment assignment, even though its effect-prediction errors are smaller because the ordering is inverted. Therefore, choosing the model with the lower (and thus better) MSE leads to a worse treatment assignment decision.

The above discussion implies that, when evaluating models for treatment assignment, what matters is their ability to minimize the expected difference between the best potential outcome that could be obtained and the potential outcome that would be obtained by using the model. This evaluation measure is also known as \textbf{expected regret (or simply ``regret'')} in decision theory:
\begin{equation}\label{eq:regret}
\text{Regret}(\hat{a}) =E[Y(a^*(X))-Y(\hat{a}(X))],
\end{equation}
where $\hat{a}$ corresponds to the evaluated treatment assignment policy, such as $\hat{a}=\textbf{1}(\hat{\tau}(x)>0)$. Regret is minimized when $\hat{a}=a^*$ (see Equation~\ref{eq:assignment_estimand}). As with causal effect estimation (discussed above), an important challenge when learning treatment assignment policies from data is that only one potential outcome is observed for each individual. However, alternative formulations of Equation~\ref{eq:regret} allow the use of historical data to evaluate models for individual treatment assignment on an aggregate basis, based on observed treatment assignments~\citep{li2010contextual,schuler2018comparison}.

So, how can machine learning methods optimize for treatment assignment itself, instead of optimizing for accurate effect estimation?
One approach to minimizing Equation~\ref{eq:regret} using machine learning is to reduce treatment assignment to importance weighted classification~\citep{zadrozny2003policy,zhao2012estimating}. As~\cite{zadrozny2003policy} first noted, treatment assignment can be framed as a weighted classification problem where the optimal treatment assignment is the target variable (as defined in Equation~\ref{eq:assignment_estimand}), and the outcomes observed under each treatment level serve to weight observations. This framing allows the use of any supervised classification algorithm to optimize for treatment assignment~\citep{beygelzimer2009offset}, but methods specifically designed for solving treatment assignment as a weighted classification problem also exist~\citep{zhao2012estimating}.

Notably, because minimizing regret is equivalent to minimizing the weighted misclassification rate as specified by~\cite{zadrozny2003policy}, a useful analogy in this context is to think about treatment assignment as a classification problem and to think about causal effect estimation as a probability estimation (or regression) problem; the two tasks are related but not the same. Importantly, it is well known in the predictive modeling literature that models with good classification performance are not necessarily good at estimating class probabilities and vice versa~\citep{friedman1997bias}. Similarly, the bias and variance components of the estimation error in causal effect predictions may combine to influence treatment assignment in a very different way than the squared error of the predictions themselves~\citep{fernandez2018causal}. So, models that are relatively bad at causal effect prediction may be good at making treatment assignments, and models that are good at causal effect prediction may be relatively bad at making treatment assignments.

Of course, in theory, the gap in performance between models optimized for causal effect estimation and models optimized for treatment assignment should disappear with unlimited data because models that accurately estimate individual causal effects should converge to optimal decision making (as shown in Equation~\ref{eq:assignment_estimand}). However, empirical evidence shows that the gap can persist even with training data consisting of more than half a billion observations~\citep{fernandez2020comparison}.\footnote{In this case, the problem was to build Spotify playlists that will lead to longer listening.} So, modeling optimal treatment assignment (rather than causal effects) can result in better causal decision making even in cases where there is plenty of data to estimate causal effects.

\subsection{Choosing the Training Data}

The second implication is that data that provide little utility for accurate causal effect estimation, due to confounding, may still be quite useful for causal decision making~\citep{fernandez2019observational}. One of the main concerns when estimating causal effects from data is the possibility of bias due to confounding, for example, due to selection bias~\citep{yahav2015tree}. Formally, if $T$ represents the treatment assignment, confounding occurs whenever $E[Y(i)|T=i,X]\neq E[Y(i)|X]$, which implies that individuals who were exposed to the treatment level $i$ are systematically different from the individuals in the overall population and, in particular, different with respect to the outcome of interest. 
Estimating causal effects from data requires that the individuals who received the treatment (the \textbf{treated}) and the individuals who did not receive the treatment (the \textbf{untreated}) be comparable in all aspects related to the outcome, except for the assigned treatment. This assumption is known as ignorability~\citep{rosenbaum1983central}, the back-door criterion~\citep{pearl2009causality}, and exogeneity~\citep{wooldridge2015introductory}. If, for example, the treated would have a better outcome than the untreated even without the treatment, then the estimation will suffer from an upward ``causal" bias\footnote{We explicitly call this ``causal" bias to distinguish it from all the other types of bias involved in statistical estimation.  In particular, this bias is different from the statistical bias usually examined in the context of machine learning's bias/variance trade-off.  A separate causal bias/variance trade-off exists~\citep{fernandez2019observational}.}; the treatment will be falsely attributed a positive causal effect that should be attributed to another unobserved, systematic difference between treated and untreated.

Bias due to (unobserved) confounding is often regarded as the kiss of death in the causal inference world.  Because it cannot be quantified from observational data alone, no additional amount of data can help if the goal is to estimate causal effects. In fact, with more data, the estimate will converge to a biased (and so, arguably wrong) estimate of the effect. Nevertheless, confounding bias does not necessarily hurt decision making~\citep{fernandez2019observational}. For example, suppose that the large prediction error made by Model 1 in Figure~\ref{fig:effect} is the result of using confounded training data to estimate the model. In this case, the confounding bias is clearly not hurtful for decision making.
In addition, by lessening potential errors due to variance, more data can potentially correct the detrimental effect of confounding bias on decision making and even confer a beneficial effect, as we illustrate below.

The implication that avoiding confounding in the data is less critical for CDM is particularly important because modern business information systems often are \emph{designed} to confound the data they produce. For example, advertisers use machine learning models to target likely buyers with ads, and websites often recommend to their users the products they are more likely to choose. If we were to use the data produced by these systems to estimate the causal effects of ads or recommendations, the estimates would almost certainly be biased because the people targeted with interventions are those who had been estimated to be likely to have a positive result (even without an intervention). Some recent developments in the estimation of treatment assignment policies from observational data are specifically motivated to deal with confounding bias~\citep{athey2021policy}. However, these approaches assume that the selection is observable (i.e., all confounding is captured by the feature vector $X$) or that treatment effects can be consistently estimated in some alternative way (e.g., by using instrumental variables). Yet evidence from large-scale experiments suggests that confounding bias is likely to persist after controlling for observable variables~\citep{gordon2019comparison}, and as pointed out above, correcting for confounding bias in causal effect estimates may actually lead to worse treatment assignments by increasing errors due to variance. 

An important alternative to dealing with confounding is to collect data for model training using randomized experiments. Estimating causal effects from observational data often is downplayed in favor of randomized experiments due to the latter’s ability to control for all confounders; the randomization procedure ensures that, on average, the treated and untreated will be comparable in all aspects other than the treatment assignment. Therefore, one could think about experiments as the natural approach to avoid confounding.  We should note that by our understanding of current practice (not academic research), most randomized experiments are not implemented to gather data for training models, but rather to evaluate specific hypotheses. This evaluation may include comparing different machine learning models, for example, to determine if a new model is statistically significantly better than the currently deployed model.  The important point here is that the amount of data needed to \textit{evaluate} model performance is usually much smaller than the amount of data needed to \textit{train} accurate models of conditional treatment effects.

This is critical because randomized experiments producing enough data to train accurate CATE models may be very expensive---in many situations prohibitively expensive---forcing us to proceed using whatever data we may have at hand. They are expensive because the treatments may be costly (especially treating those for whom the treatment would have no beneficial effect) and because of the opportunity costs incurred by withholding treatment from those for whom it would be beneficial. Moreover, randomized experiments are not always possible. For example, one cannot randomly force workers to take or switch jobs in order to assess the impact of different jobs on future career opportunities. Yet, freelancing platforms like Upwork may be interested in making job recommendations.  Business-to-business firms often cannot simply withhold services from their customers, as a consumer-oriented business might.\footnote{Understanding how confounding affects intervention decision making is useful even when experiments are feasible. For example, as we describe next, causal bias may either help or hurt decision making. Careful experimentation may help to determine whether it would be profitable to invest in correcting causal bias.}

\pgfmathdeclarefunction{gauss}{2}{%
  \pgfmathparse{1/(#2*sqrt(2*pi))*exp(-((x-#1)^2)/(2*#2^2))}%
}

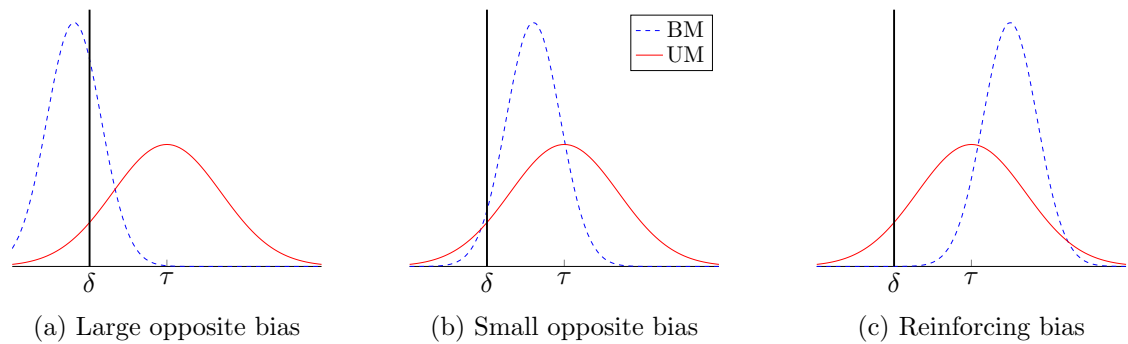
\begin{figure}
\centering
\begin{subfigure}{.32\textwidth}
  \centering
  \begin{tikzpicture}[scale=0.6]
    \begin{axis}[every axis plot post/.append style={
      mark=none,domain=-1:3,samples=50,smooth},
      axis x line*=bottom, 
      xtick={0, 1}, 
      xticklabels={$\delta$, $\tau$},
      hide y axis, ymax=1.2,
      tick label style={font=\Large},
      axis y line*=left, 
      enlargelimits=false] 
      \addplot[dashed, blue]{gauss(-0.2,0.35)};
      \addplot {gauss(1,0.7)};
      \addplot[line width=1.2pt, black] coordinates {(0,0)(0,1.2)};
    \end{axis}
    \end{tikzpicture}
  \caption{Large opposite bias}
  \label{fig:bias_high}
\end{subfigure}%
\begin{subfigure}{.32\textwidth}
  \centering
  \begin{tikzpicture}[scale=0.6]
  \begin{axis}[every axis plot post/.append style={
      mark=none,domain=-1:3,samples=50,smooth},
      axis x line*=bottom, 
      xtick={0, 1}, 
      xticklabels={$\delta$, $\tau$},
      hide y axis, ymax=1.2,
      tick label style={font=\Large},
      legend style={font=\large},
      axis y line*=left, 
      enlargelimits=false] 
      \addplot[dashed, blue]{gauss(0.6,0.35)};
      \addlegendentry{BM}
      \addplot {gauss(1,0.7)};
      \addlegendentry{UM}
      \addplot[line width=1.2pt, black] coordinates {(0,0)(0,1.2)};
   \end{axis}
   \end{tikzpicture}
  \caption{Small opposite bias}
  \label{fig:bias_med}
\end{subfigure}
\begin{subfigure}{.32\textwidth}
  \centering
  \begin{tikzpicture}[scale=0.6]
  \begin{axis}[every axis plot post/.append style={
      mark=none,domain=-1:3,samples=50,smooth},
      axis x line*=bottom, 
      xtick={0, 1}, 
      xticklabels={$\delta$, $\tau$},
      hide y axis, ymax=1.2,
      tick label style={font=\Large},
      axis y line*=left, 
      enlargelimits=false] 
      \addplot[dashed, blue]{gauss(1.5,0.35)};
      \addplot {gauss(1,0.7)};
      \addplot[line width=1.2pt, black] coordinates {(0,0)(0,1.2)};
   \end{axis}
   \end{tikzpicture}
  \caption{Reinforcing bias}
  \label{fig:bias_low}
\end{subfigure}
\vspace{3mm}
\caption{Sampling distributions of the causal effect estimate of a biased model (BM) and an unbiased model (UM). In this example (see details in the text), the correct decision is made when a model's estimate of $\tau$ is larger than $\delta$. For decision making, a BM can be either worse (a), about the same (b), or even better (c) than a UM.}
\label{fig:scenarios}
\end{figure}

Let's illustrate the feasibility of using confounded data for causal decision making with another example, shown in Figure~\ref{fig:scenarios}. In this example, we need to decide whether to intervene on a particular individual. 
Figure~\ref{fig:scenarios} shows the true causal effect $\tau$ and the optimal threshold $\delta$ for deciding whether to intervene (i.e., $\delta$ considers the costs of intervening vs. not intervening).
In our example, $\tau>\delta$, meaning that for this individual, the optimal decision is to intervene.
Of course, we do not know the true effect $\tau$, so we consider two models to estimate it: a biased model (BM) and an unbiased model (UM). Each of the panels in Figure~\ref{fig:scenarios} shows the sampling distributions of the causal effect estimates provided by the two models under different degrees of causal bias. One model (BM, in blue) is biased due to confounding that varies across panels---this bias can be seen in the figures, as the distributions of the BM's estimate are not centered on $\tau$.  At the same time, the BM benefits from smaller variance because there are plenty of naturally occurring (biased) training data.
The second model (UM, in red) is the same across all three panels, and it is unbiased---we can see that its distribution is indeed centered on $\tau$.  However, the UM suffers from larger variance because it was learned using an unconfounded training data set that is smaller due to experimentation costs. 

When using a model for decision making, we obtain an estimate drawn from the model's sampling distribution, and we intervene if the estimate of the causal effect is greater than (to the right of) $\delta$. Therefore, for each sampling distribution, the area to the left side of $\delta$ represents the probability of making the wrong decision. In this example, the UM usually leads to the correct decision (i.e., most of the red distribution falls to the right of $\delta$).\footnote{If $\tau$ had been closer to $\delta$, the UM would make the wrong decision more often.} In contrast, decision making using the BM is strongly affected by the causal bias. Figure~\ref{fig:bias_high} shows a case in which more than half of the BM distribution falls to the left of $\delta$, due to a large ``downward" bias that leads the BM to systematically underestimate $\tau$. Under these circumstances, the bias is so strong that the BM makes the wrong decision on average.  Thus, here the UM is a better alternative, despite having larger variance.

Now consider Figure~\ref{fig:bias_med}. Here the causal bias still leads the BM to systematically underestimate $\tau$, but slightly less.  This small change in effect estimation bias has a large effect on decision-making accuracy; for decision-making, the BM's smaller variance compensates for its bias.  Whereas in Figure~\ref{fig:bias_high}, the UM was a much better choice, in Figure~\ref{fig:bias_med}, it is no longer clear which model is a better alternative---the two models seem equally likely to make the correct decision. It is easy to see how the BM could be preferable despite its bias if it had even smaller variance. Importantly, in this case, both the BM and the UM would converge to the optimal decision with unlimited data because their distributions are centered on the “correct side” of the threshold, implying that the detrimental effect of confounding on decision making can be corrected with larger data. So, while the BM cannot recover the true causal effect, it can in fact identify the optimal decision. 

Finally, Figure~\ref{fig:bias_low} shows an example in which causal bias could help. In our example, this would happen when the bias leads to an overestimation of the causal effect,\footnote{More generally, it happens when the bias pushes the estimation further in the direction of the correct decision.} which lessens errors due to variance. In practice, this could occur when confounding is stronger for individuals with large effects---for example, if confounding bias is stronger for “likely buyers,” but the effect of ads is also stronger for them. In such scenarios, the BM would be a better alternative than the UM for making intervention decisions, even if its causal effect estimates are highly inaccurate.

The key insight here is that optimizing to make the correct decision generally involves understanding whether a causal effect is above or below a given threshold, which is different from optimizing to reduce the magnitude of the bias in a causal effect estimate. Hence, correcting for confounding bias when learning treatment assignment policies (e.g., as proposed in~\cite{athey2021policy}) may hurt decision making. Furthermore, models trained with confounded data may lead to decisions that are as good (or better) than the decisions made with models trained with costly experimental data~\citep{fernandez2019observational}, in particular when larger causal effects are more likely to be overestimated or when the variance reduction benefits of more and cheaper data outweigh the detrimental effect of confounding.

Although most of our discussion focuses on selection in treatment assignment as a source of confounding bias, this insight is agnostic to the source of the bias and can be generalized to other situations where the data may be problematic for causal effect estimation. For instance, the bias could be the result of violating the stable unit treatment value assumption (SUTVA), non-compliance with the treatment assignment, flaws in the experimental design, or differences between the setup in the training data and the setting where the trained model is expected to be used. The main takeaway is that data issues that make it impossible to estimate causal effects accurately do not necessarily keep us from using the data to make accurate intervention decisions. 

\subsection{Leveraging Alternative Target Variables}

The third implication that arises from distinguishing between causal decision making and causal effect estimation is that, for causal decision making, modeling with alternative (potentially non-causal) target variables may work better than predicting the actual causal effects on the outcome of interest~\citep{fernandez2018causal}.  To elaborate on this implication, consider again the binary treatment setting where we would like to treat those individuals for whom the treatment will have a positive effect.  This elaboration defines a causal classification problem: classify individuals into treat or do-not-treat.\footnote{The binary treatment distinction is purely for ease of illustration.  With multiple treatment options, we have a multiclass classification problem.  As mentioned earlier, causal decision making can be reduced to a classification task where individuals should be classified into the treatment level (class) that produces the most beneficial outcome~\citep{zadrozny2003policy,zhao2012estimating}.}  Viewing CDM as a classification task allows us to draw an analogy with the distinctions between classification versus regression and classification versus probability estimation that have been studied in depth in non-causal predictive modeling.  Specifically, many classification decisions essentially involve comparing an instance's estimated class membership probability to a threshold and then taking action based on whether the estimate is above or below the threshold.\footnote{The threshold could be based on a cost/benefit matrix~\citep{provost2013data}, a budget or workforce constraint~\citep{provost2001robust}, or even instance specific characteristics~\citep{saar2007decision}.}
In such situations, causal effect estimates are relevant for decision making only to the extent that they help to discriminate between the desired actions, for example, those who would benefit from being treated versus those who would not.  

What does this have to do with alternative target variables?  For traditional classification, we know that often we can build more effective predictive models by using a proxy target variable, rather than a target variable that represents the true outcome of interest~\citep{provost2013data}.  
Proxy modeling is often applied when data on the true objective are in short supply (or even completely nonexistent).  
For example, in online advertising, often few data are available on the ultimate desired conversion (e.g., purchases) due to low conversion rates or cold start periods. This can pose a challenge for accurate campaign optimization. In this context, \cite{dalessandro2015evaluating} demonstrate the effectiveness of proxy modeling and find that predictive models built based on brand site visits (which are much more common than purchases) do a remarkably good job of predicting which online consumers will purchase.  Causal modeling also often suffers from scant training data; similarly to non-causal predictive modeling, estimating the best action based on a proxy target may work better than directly estimating causal effects on the outcome of interest. 

The use of proxy outcomes to build predictive conditional causal models should not be a major paradigm shift for researchers, because proxy or ``surrogate" outcomes already are sometimes used to evaluate treatment effects~\citep{prentice1989surrogate,vanderweele2013surrogate}. For example, in clinical trials, the goal is often to study the efficacy of an intervention on outcomes such as patients’ long-term health or survival rate. However, the primary outcome of interest might be rare or only observed after years of delay (e.g., a 5-year survival rate). In such cases, it is common to use the effect of an intervention on surrogate outcomes as a proxy for its effect on long-term outcomes. Recently, \cite{yang2020targeting} propose how to use surrogate outcomes to impute the missing long-term outcomes and use the imputed long-term outcomes to optimize a targeting policy; they demonstrate their approach in the context of proactive churn management. The use of short-term proxies to estimate long-term treatment effects is discussed in more detail by~\cite{athey2016estimating}.

In practice, it is common for firms to make targeted intervention decisions without any causal modeling at all. Instead, in large-scale predictive applications such as ad targeting and churn incentive targeting, practitioners typically target based on models predicting outcomes (e.g., whether someone will buy after being shown an ad) rather than models predicting treatment effects~\citep{ascarza2018pursuit}.  For example, in targeted online advertising, although ad campaigns are sometimes evaluated in terms of their causal effect (e.g., through A/B testing), causal effect models are rarely used for actually targeting the ads. Instead, potential customers are usually targeted based on predictions of how likely they are to convert, which are biased estimates of individual causal effects because they do not account for the counterfactual (e.g., a customer may buy even without the ad). Researchers and savvy practitioners see this as due either to naivety, lack of modeling sophistication, or pragmatics~\citep{ascarza2018pursuit, provost2013data}. However, outcome prediction models are used even when practitioners have looked carefully at the estimation of treatment effects (cf.,~\cite{stitelman2011estimating} and~\cite{perlich2014machine} for a clear example).  We suggest that this use of outcome models rather than causal-effect models can be seen as proxy modeling---and we should not simply criticize the practitioners for lack of modeling sophistication, but rather ask what George Box might have asked: are the proxy models more useful?

Although practitioners should indeed proceed with caution when targeting with outcome prediction models~\citep{ascarza2018retention,radcliffe2011real}, prior research studies do show evidence that targeting based on outcome predictions can be effective. For example, in an advertising campaign of a major food chain, \cite{stitelman2011estimating} found that displaying ads based on likelihood of conversion increased the volume of prospective new customers by 40\% more than displaying ads to the general population. \cite{huang2015telco} deployed a churn prediction model in a telecommunications firm that increased the recharge rate of potential churners by more than 50\% according to an A/B test. \cite{mackenzie2013retailers} estimated that 35\% of what consumers purchase on Amazon and 75\% of what they watch on Netflix come from product recommendations based on non-causal predictive models (as we understand it). None of these targeting approaches modeled more than one potential outcome, so it would be wrong\footnote{Except under extreme assumptions.} to interpret the model estimates as causal effects; however, their estimates were able to identify individuals who would be positively affected by interventions.

Consistent with the proxy-modeling view, using outcome predictions can result in more effective causal decision making than using causal effect predictions. \cite{radcliffe2011real} provide a list of several conditions under which causal effect predictions may not produce better targeting results than outcome predictions, such as when treatment effects are particularly small, and outcomes are informative of effects. ~\cite{fernandez2018causal} explain why, presenting a theoretical analysis comparing the two approaches and showing that larger sampling variance in treatment effect estimation may lead to more decision-making errors than the (biased) outcome prediction approach. Moreover, they show that the often stronger signal-to-noise ratio in outcome prediction may help with intervention decisions when outcomes and treatment effects are correlated. Thus, it may be possible to make good targeting decisions in some settings without actually having any data on how people behave when targeted.

We illustrate this last point using data made available by Criteo (an advertising platform) based on randomly targeting advertising to a large sample of users~\citep{diemert2018large}.\footnote{Check \URL{https://ailab.criteo.com/criteo-uplift-prediction-dataset/} for details and access to the data. We use the version of the dataset without leakage.} We use “conversions” as the outcome of interest.  The data include 13,979,592 instances, each representing a user with 11 features, a treatment indicator for the ad, and the label (i.e., whether the user converted or not). The treatment rate is 85\%.
\begin{figure}
    \centering
    \includegraphics[width=0.7\textwidth]{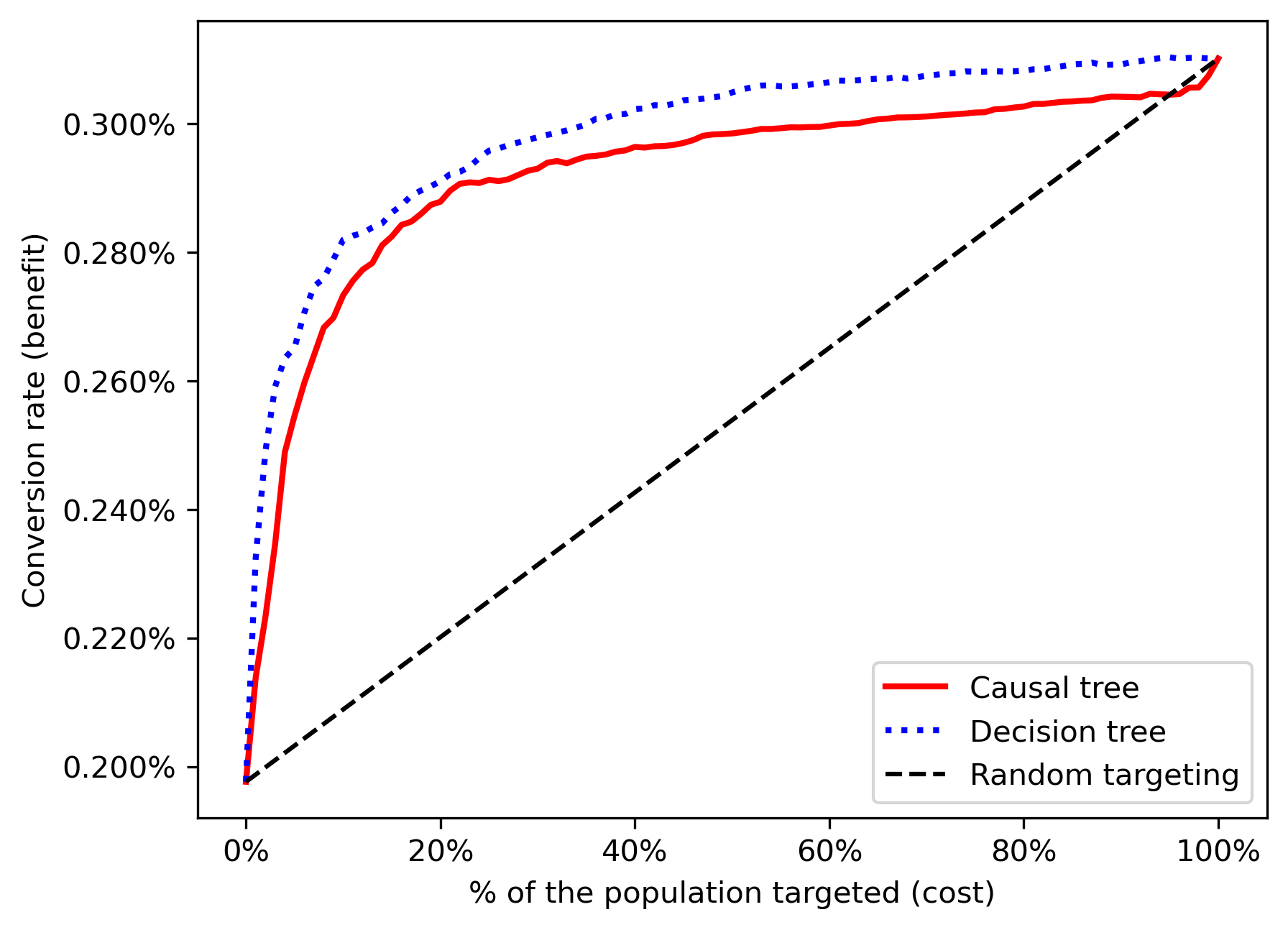}
    \caption{Increase in conversions produced by an outcome model and an uplift model. In this specific example, the non-causal model can produce at least as much benefit as the causal model for any given cost.}
    \label{fig:comparison}
\end{figure}

We consider two different targeting models: a causal tree~\citep{athey2016recursive} that predicts the CATE of the ad on conversions, and a non-causal decision tree that predicts the probability of conversion when no ad is shown. Both models were trained and tuned with cross-validation using 80\% of the sample (the training set), but we only used the data for the untreated instances to train and tune the non-causal tree. So, the non-causal tree was trained with no data on how people behave when treated and substantially fewer data overall (recall that the treatment rate is 85\%). 

We then evaluated the two models on the remaining 20\% of the sample (the test set) by using their predictions to score individuals and plotting the expected conversion rate as a function of the percentage of the individuals with the largest scores who are targeted, as shown in~Figure~\ref{fig:comparison}. More specifically, because the data was collected through a randomized experiment, we can estimate the conversion rate produced by a targeting model as:
\begin{equation}\label{eq:evaluation}
    E[D]
    \times E[Y|T=1,D=1] + (1-E[D])\times E[Y|T=0,D=0],
\end{equation}
where $D$ is a binary variable that indicates whether the evaluated model intends to target the individual, $T$ is a binary variable that indicates whether the individual was treated in the data, and $Y$ is a binary variable that indicates whether the individual converted. So, Figure~\ref{fig:comparison} estimates the conversion rates that would result from using the evaluated models to decide which individuals to target, as the models target more individuals. 

Surprisingly, the (non-causal) decision tree is better than the causal tree at identifying the candidates for whom the ad is most effective. The explanation behind this counterintuitive result is that features predictive of conversions are also predictive of treatment effects. However, the signal-to-noise ratio is higher for conversions than for treatment effects. Therefore, it is easier for machine learning algorithms to discriminate individuals according to outcomes than according to treatment effects. Critically, the training data set used to estimate the causal tree is more than six times larger than the training data set used to estimate the decision tree. Thus, even though (in theory) the causal tree should outperform the decision tree with more data, acquiring enough training data to reach that point may be overly expensive in this case (because the data is collected by randomly targeting and withholding treatments).

That said, outcome prediction models also have been shown to perform poorly in some cases, such as in customer retention applications where modeling negative effects is important~\citep{ascarza2018retention,radcliffe2011real}. 

Generally, outcome prediction can be a reasonable alternative to causal effect estimation when~\citep{fernandez2018causal}:
\begin{enumerate}
\item Outcomes and effects are correlated. For instance,~\cite{radcliffe2011real} find that marketing interventions often have the most positive effect on high-spending customers, and as a result, modeling causal effects does not benefit decision making as much in retail environments. 
\item Outcomes are easier to estimate than causal effects, which may occur when data for one or more treatment conditions are limited (e.g., due to experimentation costs) or the signal-to-noise ratio is much larger for outcomes than for effects (as in our example). 
\item Predictions are used to rank individuals. For example, if the goal is to intervene on the top-percent individuals with the largest effects, we only need to score and rank the individuals; whether the scores represent effect estimates is irrelevant.  
\end{enumerate}

The conditions in this list are interrelated. For example, when outcomes are much easier to estimate than causal effects, a moderate correlation between outcomes and effects could be enough for an outcome prediction model to outperform a causal effect model. Similarly, a high correlation between outcomes and effects implies that ranking individuals according to outcomes or causal effects would yield similar results, so the ability to estimate outcomes better than causal effects is not as relevant. These observations apply to the specific case where the models are used to rank and then intervene on individuals (the third condition), so whether outcome prediction models could be useful for other causal tasks is an open question.

Outcome prediction is only one possible proxy task for CDM, but it is by far the most commonly employed (consciously or not).
For additional practical advice on when to deploy a causal effect prediction model instead of an outcome prediction model, see \cite{radcliffe2011real}.

\section{Conclusions and Future Research}

This paper develops the perspective that causal decision making and causal effect estimation are fundamentally different tasks.  They are different tasks because their estimands of interest are different. Causal effect estimation assesses the impact of an intervention on an outcome of interest, and often we want or need accurate (and unbiased) estimations.  In contrast, unbiased and accurate estimation of causal effects is not necessary for causal decision making, where the goal is to decide what treatment is best for each individual---a causal classification problem.

We present implications of making this distinction on (1) choosing the algorithm used to learn treatment assignment policies, that is, causal predictive models, (2) choosing the training data used to train the causal models, and (3) choosing the target variable for training the causal models. Importantly, we highlight that traditionally ``good'' estimates of causal effects are not necessary to make good causal decisions.

In a recent special issue of \textit{Observational Studies}~\citep{mitra2021introduction}, thought leaders from across the globe wrote comments discussing Leo Breiman's influential paper ``Statistical Modeling: The Two Cultures''~\citep{breiman2001statistical}. In his visionary paper, Breiman argued for an algorithmic modeling culture that should validate statistical models based on their prediction accuracy. When that paper was published, some individuals deemed this prediction-based focus incompatible with causal modeling~\citep{cox2001statistical}, but new commentaries suggest that this perspective is not at odds when ``\emph{guided by a formal causal model for identification and bias reduction}''\citep{pearl2021causally} and the algorithms incorporate ``\emph{economic causal restrictions and non-prediction objectives}''~\citep{imbens2021breiman}.

We take a step further and argue that defining an appropriate prediction accuracy (evaluation) measure is perhaps even more important to integrate causal modeling with an algorithmic modeling culture. Both CEE and CDM can be, after all, predictive tasks. CEE can be used to predict what would happen under a counterfactual scenario, whereas CDM consists of predicting the best (counterfactual) course of action. The models can have a good CDM performance even when traditional causal assumptions are violated  or not incorporated by the models.

Of course, this does not imply that firms should stop investing in randomized experiments or that causal effect estimation is not relevant for decision making. The argument here is that causal effect estimation is not necessary for \emph{doing} effective treatment assignment. However, causal effect estimation is certainly important for the \emph{evaluation} of causal decisions, particularly if there are concerns about how and why the decisions were made. Therefore, this paper should not be interpreted as a discourse against randomized experiments or methods for causal effect estimation. We hope this paper will encourage practitioners to run randomized experiments evaluating all sorts of treatment assignment policies, including policies based on non-causal statistical models. Furthermore, randomized experiments can be employed to identify the underlying theoretical mechanisms that result in some treatments being more effective for certain subpopulations~\citep{tafti2020beyond}, which may lead to the development of even more effective treatments.

There are multiple major opportunities for future research in causal decision making. First, most studies that distinguish between causal decision making and causal effect estimation do so in the context of decisions that are categorical and independent. This scope limitation has allowed researchers to draw parallels between causal decision making and classification. However, few studies have discussed continuous decisions~\citep{dube2019personalized} or settings where multiple treatments are assigned to the same individuals over time~\citep{chakraborty2014dynamic}. The results discussed in this paper imply that what matters in causal decision making is the \emph{ranking} of the alternatives available to the decision maker rather than accurately estimating the outcomes or effects associated with each alternative. Therefore, prediction errors that preserve the ranking should not affect continuous or time-dependent decisions either. A unified and general framework explaining how prediction errors may affect causal decision making in broader settings is thus a natural step for future research. 

Another promising direction for future work is the mathematical formalization of the conditions under which optimal treatment assignments can be learned from data. In causal effect estimation, several studies have proposed mathematical assumptions to characterize the conditions under which causal effects can be consistently estimated from data. Well-known examples include SUTVA~\citep{cox1958planning}, ignorability~\citep{rosenbaum1983central}, back-door criterion~\citep{pearl2009causality}, front-door criterion~\citep{pearl2009causality}, transportability~\citep{pearl2011transportability}, and other assumptions for the identification of causal effects using instrumental variables~\citep{angrist1996identification}. The perspective presented in this paper suggests that such assumptions are not necessary to learn optimal treatment assignments from data. As an example, assuming the unconfoundedness assumption is violated and the selection mechanism producing the confounding is a function of the causal effect---so that the larger the causal effect the stronger the selection---then (intuitively) the ranking of the preferred treatment alternatives should be preserved in the confounded setting, allowing the estimation of optimal treatment assignment policies from data. Mathematically formalizing such settings and linking them to real-world applications is a promising direction for future research.

The results discussed here also suggest that machine learning procedures that combine multiple sources of data (e.g., observational and experimental data) could conceivably lead to better intervention decisions by optimizing the trade-off between confounding bias and supervised learning errors. Hence, another promising area for future research in causal decision making is the development of methods that can combine multiple sources of data.
Several pioneering efforts to combine observational and experimental data for causal inference have been made~\citep{peysakhovich2016combining,rosenman2020combining,kallus2018removing,fernandez2020combining,athey2020combining}, but most have the intention of estimating causal effects (rather than treatment assignment policies). Another interesting direction for the development of new methods is to try ``clever'' experimentation. For instance, decision makers could focus on running randomized experiments in parts of the feature space where the confounding is particularly hurtful for decision making, resulting in higher returns on their experimentation budget. Recent efforts seeking to maximize the decision-making benefits of randomized controlled trials while minimizing experimentation costs~\citep{feit2019test,simester2020efficiently,gangarapu2020multi,miller2020test} suggest that this is a particularly promising research direction.

\bibliographystyle{apalike}
\bibliography{commentary}

\end{document}